\documentclass[sigconf]{acmart}

\usepackage{amsmath,amsfonts,bm}

\def\eqref#1{equation~\ref{#1}}

\def\1{\bm{1}}

\DeclareMathAlphabet{\mathsfit}{\encodingdefault}{\sfdefault}{m}{sl}
\SetMathAlphabet{\mathsfit}{bold}{\encodingdefault}{\sfdefault}{bx}{n}

\usepackage{todonotes}

\newcommand{\SC}{\textit{Starcraft-2~}}

\newcommand{\editcolor}{black}
\newcommand{\newedit}[1]{\textcolor{\editcolor}{#1}}

\usepackage{float}
\usepackage{array,multirow}
\usepackage{hyperref}
\usepackage{wrapfig}
\hypersetup{
    colorlinks=true,
    linkcolor=red,
    filecolor=magenta,      
    urlcolor=blue,
    citecolor=purple,
    pdftitle={Overleaf Example},
    pdfpagemode=FullScreen,
    }
\usepackage{graphicx}
\usepackage{algorithm}
\usepackage{algorithmic}
\usepackage{lineno}
\usepackage{color}
\usepackage[normalem]{ulem}%
\usepackage{soul}

\usepackage{balance}

\usepackage[compact]{titlesec}
\titlespacing*{\section}{0pt}{*0}{*0}
\titlespacing*{\subsection}{0pt}{*0}{*0}
\titlespacing*{\subsubsection}{0pt}{*0}{*0}

\AtBeginDocument{%
  \providecommand\BibTeX{{%
    \normalfont B\kern-0.5em{\scshape i\kern-0.25em b}\kern-0.8em\TeX}}}

\copyrightyear{2022}
\acmYear{2022}
\setcopyright{acmcopyright}
\acmConference[AIMLSystems 2022]{The Second
International Conference on AI-ML Systems}{October 12--15, 2022}{Bangalore,
India}
\acmBooktitle{The Second International Conference on AI-ML Systems
(AIMLSystems 2022), October 12--15, 2022, Bangalore, India}
\acmPrice{15.00}
\acmDOI{10.1145/3564121.3565236}
\acmISBN{978-1-4503-9847-3/22/10}

\begin{document}
\title{System Design for an Integrated Lifelong Reinforcement Learning Agent for Real-Time Strategy Games
}
\renewcommand{\shorttitle}{System Design for an Integrated L2 RL Agent for Real-Time Strategy Games}

        \author{Indranil Sur}
        \authornote{Equal contribution}
        \authornote{SRI International, Princeton, NJ, USA}
        \orcid{0000-0002-0135-4463}
        \affiliation{%
            \city{Princeton}
            \state{NJ}
            \country{USA}
        }
        \email{indranil.sur@sri.com}

        \author{Zachary Daniels}
        \authornotemark[1]
        \orcid{0000-0002-6819-4599}
        \authornotemark[2]
        \affiliation{%
            \city{Princeton}
            \state{NJ}
            \country{USA}
        }
        \email{zachary.daniels@sri.com}

        \author{Abrar Rahman}
        \orcid{0000-0003-4937-5500}
        \authornotemark[2]
        \affiliation{%
            \city{Princeton}
            \state{NJ}
            \country{USA}
        }
        \email{abrar.rahman@sri.com}

        \author{Kamil Faber}
        \orcid{0000-0003-4221-0017}
        \authornote{American University, Washington, DC, USA}
        \affiliation{%
            \city{Washington}
            \state{DC}
            \country{USA}}
        \email{kfaber@agh.edu.pl}

        \author{Gianmarco J. Gallardo}
        \orcid{0000-0001-7549-2917}
        \authornote{Rochester Institute of Technology, Rochester, NY}
        \affiliation{%
        \city{Rochester}
        \state{NY}
        \country{USA}}
        \email{gg4099@rit.edu}
        
        \author{Tyler L. Hayes}
        \orcid{0000-0002-0875-7994}
        \authornotemark[4]
        \affiliation{%
            \city{Rochester}
            \state{NY}
            \country{USA}}
        \email{tlh6792@rit.edu}

        \author{Cameron E. Taylor}
        \orcid{0000-0002-1345-6563}
        \authornote{Georgia Institute of Technology, Atlanta, GA, USA}
        \affiliation{%
            \city{Atlanta}
            \state{GA}
            \country{USA}}
        \email{cameron.taylor@gatech.edu}

        \author{Mustafa Burak Gurbuz}
        \orcid{0000-0001-8267-7821}
        \authornotemark[5]
        \affiliation{%
            \city{Atlanta}
            \state{GA}
            \country{USA}}
        \email{mgurbuz6@gatech.edu}

        \author{James Smith}
        \orcid{0000-0001-9210-0161}
        \authornotemark[5]
        \affiliation{%
            \city{Atlanta}
            \state{GA}
            \country{USA}}
        \email{jamessealesmith@gatech.edu}

        \author{Sahana Joshi}
        \orcid{0000-0003-2741-2948}
        \authornotemark[5]
        \affiliation{%
            \city{Atlanta}
            \state{GA}
            \country{USA}}
        \email{sjoshi330@gatech.edu}

        \author{Nathalie Japkowicz}
        \orcid{0000-0003-1176-1617}
        \authornotemark[3]
        \affiliation{%
            \city{Washington}
            \state{DC}
            \country{USA}}
        \email{japkowic@american.edu}

        \author{Michael Baron}
        \orcid{0000-0003-2759-1181}
        \authornotemark[3]
        \affiliation{%
            \city{Washington}
            \state{DC}
            \country{USA}}
        \email{baron@american.edu}

        \author{Zsolt Kira}
        \orcid{0000-0002-2626-2004}
        \authornotemark[5]
        \affiliation{%
            \city{Atlanta}
            \state{GA}
            \country{USA}}
        \email{zkira@gatech.edu }
        
        \author{Christopher Kanan}
        \orcid{0000-0002-6412-995X}
        \authornote{University of Rochester, Rochester, NY}
        \affiliation{%
            \city{Rochester}
            \state{NY}
            \country{USA}}
        \email{ckanan@cs.rochester.edu}
        
        \author{Roberto Corizzo}
        \orcid{0000-0001-8366-6059}
        \authornotemark[3]
        \affiliation{%
            \city{Washington}
            \state{DC}
            \country{USA}}
        \email{rcorizzo@american.edu}
        
        \author{Ajay Divakaran}
        \orcid{0000-0003-0371-5346}
        \authornotemark[2]
        \affiliation{%
            \city{Princeton}
            \state{NJ}
            \country{USA}
        }
        \email{ajay.divakaran@sri.com}

        \author{Michael Piacentino}
        \orcid{0000-0001-9954-606X}
        \authornotemark[2]
        \affiliation{%
            \city{Princeton}
            \state{NJ}
            \country{USA}
        }
        \email{michael.piacentino@sri.com}

        \author{Jesse Hostetler}
        \orcid{0000-0002-1757-7493}
        \authornotemark[2]
        \affiliation{%
            \city{Princeton}
            \state{NJ}
            \country{USA}
        }
        \email{jesse.hostetler@sri.com}

        \author{Aswin Raghavan}
        \orcid{0000-0001-5052-1929}
        \authornotemark[2]
        \affiliation{%
            \city{Princeton}
            \state{NJ}
            \country{USA}
        }
        \email{aswin.raghavan@sri.com}

\renewcommand{\shortauthors}{Sur and Daniels, et al.}

\begin{abstract}

    As Artificial and Robotic Systems are increasingly deployed and relied upon for real-world applications, it is important that they exhibit the ability to continually learn and adapt in dynamically-changing environments, becoming \emph{Lifelong Learning Machines}. Continual/lifelong learning (LL) involves minimizing catastrophic forgetting of old tasks while maximizing a model's capability to learn new tasks. This paper addresses the challenging lifelong reinforcement learning (L2RL) setting. Pushing the state-of-the-art forward in L2RL and making L2RL useful for practical applications requires more than developing individual L2RL algorithms; it requires making progress at the systems-level, especially research into the non-trivial problem of how to integrate multiple L2RL algorithms into a common framework. In this paper, we introduce the \emph{Lifelong Reinforcement Learning Components Framework (L2RLCF)}, which standardizes L2RL systems and assimilates different continual learning components (each addressing different aspects of the lifelong learning problem) into a unified system. As an instantiation of L2RLCF, we develop a standard API allowing easy integration of novel lifelong learning components. We describe a case study that demonstrates how multiple independently-developed LL components can be integrated into a single realized system. We also introduce an evaluation environment in order to measure the effect of combining various system components. Our evaluation environment employs different LL scenarios (sequences of tasks) consisting of \SC minigames and allows for the fair, comprehensive, and quantitative comparison of different combinations of components within a challenging common evaluation environment.

\end{abstract}

\ccsdesc[500]{Computing methodologies~Machine learning}
\ccsdesc[500]{Computing methodologies~Reinforcement learning}
\ccsdesc[500]{Computing methodologies~Artificial intelligence}
\ccsdesc[500]{Information systems}

\keywords{
  Lifelong Learning, 
  Reinforcement Learning, 
  System Design,
  Integrative Component Framework,
  \SC
}


\maketitle
\begin{figure*}[h]
  \includegraphics[width=\textwidth]{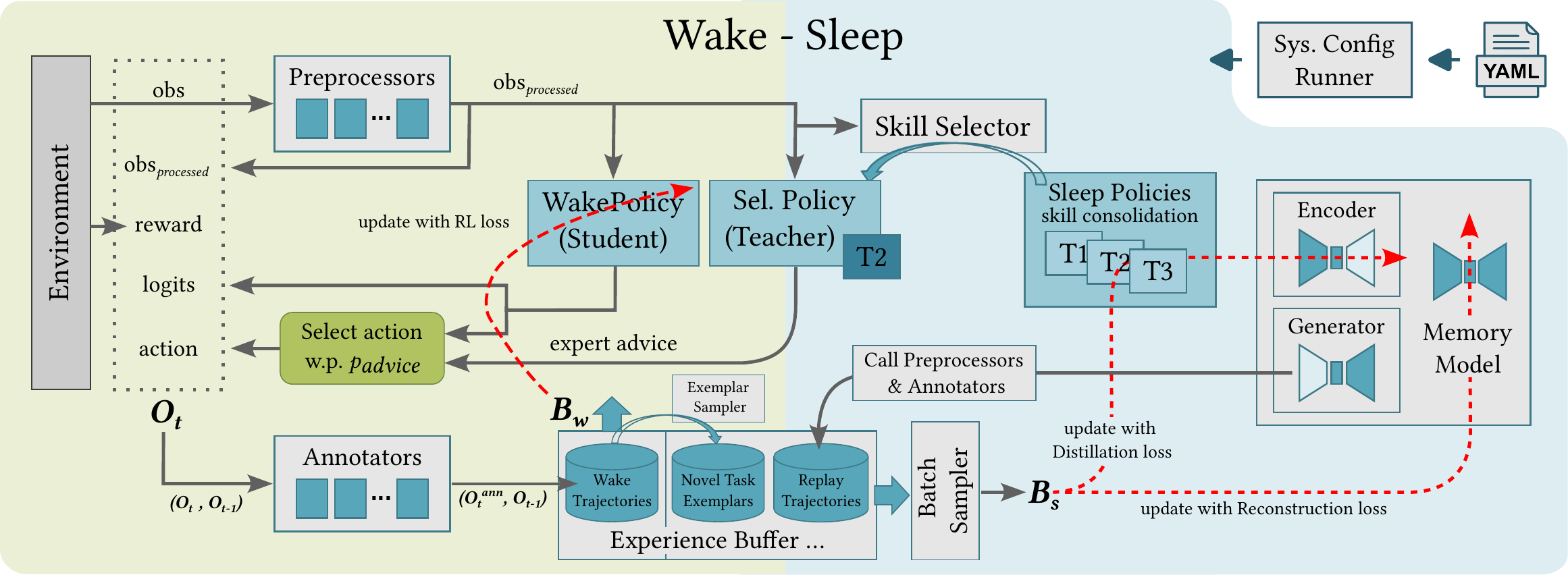}
  \caption{Lifelong RL Components Framework: The figure delineates the inner workings of a highly configurable, modular, extendable framework. The framework is needed to standardize Lifelong Reinforcement Learning (L2RL) systems and assimilate different continual learning ideas into a unified system.}
  \label{fig:main}
\end{figure*}

\section{Introduction}



Machine learning-based Artificial and Robotic Systems generally follow the paradigm of training once on a large set of data after which they are deployed and rarely updated. Improving these systems as additional training data is collected or for adaptation to new tasks requires expensive, offline fine-tuning or re-training. In contrast, humans and animals continue to learn new concepts and evolve their skill sets as they act within and interact with novel environments over long lifespans. That is, biological systems demonstrate the ability to continuously acquire, fine-tune, and adequately reuse skills in novel combinations to solve novel yet structurally-related problems \cite{mendez2020lifelong}. As Artificial and Robotic Systems are increasingly relied upon for mission-critical real-world applications, it is increasingly important that they exhibit similar capabilities and are able to continually learn and adapt in dynamically-changing environments, truly becoming \emph{Lifelong Learning Machines}. 


Continual learning and lifelong learning (L2) \cite{thrun1995lifelong} remain long-standing challenges for the machine learning community \cite{parisi_continual_2019}. Even for simple classification tasks, i.e., incrementally learning to classify a new category of object, leads to \textit{catastrophic forgetting} \cite{kirkpatrick_overcoming_2016}. On the other end, models may exhibit too much rigidity during learning by prioritizing preserving performance on old classes/tasks and failing to learn the new classes/tasks.
Minimizing catastrophic forgetting while maximizing the capability to learn new tasks is a key issue faced when designing lifelong learning algorithms. Balancing these two properties is known as the \textit{the Stability-Plasticity Dilemma} \cite{parisi_continual_2019}. 

There has been a lot of progress on continual learning for the incremental classification task \cite{de2021continual,mai2022online}, but continual learning in a reinforcement learning setting is more challenging
\cite{khetarpal2020towards,padakandla2021survey}, and research on lifelong reinforcement learning (L2RL) is still in its infancy. Due to the interactive nature of the setting, pushing the state-of-the-art forward in L2RL and making L2RL useful for real-world applications requires going beyond researching and developing L2RL algorithms. It also requires careful investigation and research at the systems level, especially research into the appropriate methods for the non-trivial integration of multiple L2RL algorithms and components into a common framework. Hence, there is a need for a highly configurable, modular, and extendable framework targeting the L2RL domain.
In recent years, there has been a push to develop systems to facilitate research in L2RL. Frameworks developed in \cite{powers2021cora}, \cite{johnson2022l2explorer}, \cite{fendley2022continual} have looked at standardizing and benchmarking L2RL scenarios and have standardized metrics for comparing L2RL algorithm performance.
\newedit{This paper addresses the unexplored challenge of combining multiple continual learning components in a common framework. We take inspiration from Complementary Learning Systems, Parisi et al. \cite{parisi_continual_2019} and assimilate different continual learning components (each addressing different aspects of the L2RL problem).}

The socio-economic impact of such a system is enormous. 
We highlight a few examples. 
i) Autonomous vehicles \cite{ma2020artificial} should adapt to changing conditions (e.g., weather, lighting) and should learn from their mistakes (e.g., accidents) in order to improve in terms of safety and utility over time. 
ii) caregiver/companion robots \cite{wikipedia_2022} should learn to adapt to the needs of specific human patients/partners. 
iii) Systems for medical diagnosis and treatment planning need to adapt to novel conditions (e.g., new disease variants) as well as adapt to the current state of patients and their response to previous interventions. 
iv) Network security systems must be able to protect against novel threats (e.g., new viruses, hacking efforts) in an expedient manner in order to minimize security breaches.
Systems exist in these application domains, but often rely on brute-force approaches (learning once from massive data) and are not truly solving the core problem. Hence, more focused efforts towards building \emph{Lifelong Learning Machines} are needed. 
v) Model obsolescence \citep{sculley2015hidden, alahdab2019empirical, bogner2021characterizing} of machine learning-based software systems is a major problem facing the software industry. 
The inherent plastic-yet-stable nature of lifelong learners enables these systems to be more robust to obsolescence resulting from data drift, concept drift, and task changes. 
Designing Artificial and Robotic Systems with lifelong learning at their core will ultimately decrease system downtime and reduce the overhead of model re-training. The technical contributions of this paper include the following:
\begin{enumerate}
   \item We describe a \emph{Lifelong Reinforcement Learning Components Framework} \emph{(L2RLCF)} that leverages a \emph{Wake-Sleep Mechanism} \cite{raghavan2020lifelong} from Complementary Learning Systems. 
   \item As an instantiation of L2RLCF, we develop a Python API allowing easy integration of novel lifelong learning components with each component encapsulated in its own module. We discuss the details of our API in Section \ref{sec:system}.
   \item  We introduce an evaluation environment to allow for the fair, comprehensive, and quantitative comparison of different combinations of components within a common evaluation environment. This environment employs different lifelong learning scenarios (sequences of tasks) consisting of \SC minigames, a challenging environment for L2RL (Section \ref{sec:evaluation_env}).
\item We demonstrate the utility of L2RLCF in a case study integrating a diverse set of independently-developed L2 algorithms from recent work in continual learning (Section \ref{sec:case_study}). We integrate: i) a base system for L2RL in a wake-sleep setting, ii) automatic triggering of sleep via change point detection \cite{faber2021watch}, iii) compression of experiences in the replay buffer \cite{hayes2019remind}, iv) task-specific prioritized replay, and v) representation learning via self-supervised learning.

\end{enumerate}

\section{L2RL Components Framework}
\label{sec:system}

In this section, we discuss the design of the Lifelong RL Components Framework (L2RLCF). A visual representation is seen in Figure \ref{fig:main}, and an algorithmic overview of the system is seen in Algorithm \ref{alg:wakesleep}.

\subsection{Wake-Sleep Mechanism}
Our framework employs a wake-sleep learning paradigm (fast and slow, respectively). Wake-sleep is a biologically-motivated framework that directly tackles the tradeoff in lifelong learning between \emph{plasticity}, i.e.\! learning the current task, and \emph{stability}, i.e.\! remembering past tasks. It was first introduced in \cite{raghavan2020lifelong} for class-incremental learning and extended to L2RL in \cite{daniels2022model}. It consists of two phases: i) a wake phase where standard non-lifelong learners learn aspects of the current task, e.g.\! off-the-shelf RL trained on the current task (current MDP reward and dynamics), and ii) a sleep phase where knowledge is consolidated across multiple wake periods. Additional details about our specific L2RL wake-sleep implementation can be found in Section \ref{sec:wakesleep}. Wake-sleep is well suited as the core for a unifying framework for any lifelong learning system architecture because it is easily extendable, modularized, and compatible with many existing continual learning algorithms.


\subsection{System Design}

\subsubsection{Environment}\hfill\\
In this paper, the environment can be any Partially Observable Markov Decision Process (POMDP) where an agent can perceive observations, interact with the environment, and receive a reward for the actions taken. In this paper, our environment is  the complex strategy game of \SC using the PySC2 \cite{vinyals2017starcraft} interface to the game engine. \SC mini-games require high sample complexity for single-task RL, and the issue is exacerbated when a sequence of POMDPs are presented. \SC requires processing percepts and requires decision-making over a large action space with multiple agents. More details about our \SC environment appear in Section \ref{sec:evaluation_env}.

To instantiate the sequence of tasks for the L2RL agent, we implement a ``Syllabus Runner'' that takes a configuration of task orderings with the length of each learning and evaluation period and automatically generates a sequence of simulators (parallelizing individual episodes within each learning phase). 

\subsubsection{Preprocessors \& Annotators}\hfill\\
There are several base classes that continual learning algorithms can implement. In this section, we focus on two of them. The \emph{Preprocessors} class consists of a list of preprocessor objects from different components. Each preprocessor comes into play when the agent perceives an observation. Each preprocessor is passed the observation features output by the simulator and subsequently transforms them in a meaningful way (e.g., converting RGB features to features usable by later machine learning algorithms). Once the observations are preprocessed, they are added to the original observations as named tuples and later system components can utilize these preprocessed features as needed.

The \emph{Annotators} class consists of annotator objects from different components. It is triggered after an agent has stepped through an action and received a reward. The tuple of (observation at previous time step $obs_{processed}$, wake policy logits, action, reward) is passed to each annotator object, which is then queried for the annotation feature. Annotaters also have access to previous observations; this can be leveraged to add interesting functionality to the system, e.g., for creating prioritized replay buffers. In the \SC case study, we used an annotator for ``danger detection'', i.e., scoring the estimated level of danger of a given state, and then building a replay buffer of safe states to promote a useful bias (avoid bad terminal states) in the policy. Like the Preprocessors class, the annotator features are added as a named tuple to get $O_{ann}$, which is then passed around in the system and used as needed.




\subsubsection{Memory Model}\hfill\\
The \emph{Memory Model} class represents generic memory models. In our experiments, this entailed different styles of replay (experience replay, generative replay), but it is flexible to more complex models such as clustering-based hierarchical memory models. There are two components within the memory model: the \emph{Encoder} and the \emph{Decoder}. The Encoder is not a necessity as some memory models will not require encoding (e.g., experience replay). The API is expressive enough to define whether the policy networks are built on top of an observation space, a processed feature space from the Preprocessors, a combination of features, or even on top of a generative model's encoder. 
The Decoder is equally flexible. It can be used in conjunction with genarative models to sample novel experiences (e.g.\!, we used a variational autoencoder as our generative memory model). In cases where generative memory is not necessary, it provides a mechanism to sample of old experiences stored in a buffer (for example, by returning exemplars).

        
        

        

\subsubsection{Wake Learner}\hfill\\
The model has a \emph{Wake Learner} that is any standalone off-the-shelf RL learner. We used Vtrace \cite{espeholt2018impala} in our experiments.

\subsubsection{Sleep Learner \& Skill Consolidation}\hfill\\
The \emph{Sleep Learner} is a different instantiation of the same RL learner as the Wake Learner and serves as a means of consolidating skills across multiple wake-sleep cycles. The sleep model is built around the idea of replaying of old experience (past samples, exemplars, generated samples, etc.) to help minimize catastrophic forgetting and learn to generalize to unseen tasks. It should be noted that in the Sleep Learner, we can model one or more sub-policies, enabling useful features such as selecting different policies based on perceptual/semantic similarity between the current and past tasks (implemented as annotator since task boundaries are unknown).  

\subsubsection{Experience Buffer}\hfill\\
The \emph{Experience Buffer} can take several forms and involve several different sampling mechanisms, including:
    \begin{itemize}
        \item \textbf{Wake Buffer}: The current interactions after passing through annotators and preprocessors is saved to wake buffer. The observations are kept sequentially to create trajectories. After the buffer becomes full, it is sampled and used for training the Wake Learner.
        \item \textbf{Exemplar Sampler}: (Optional) We have observed if certain replay architectures are used (e.g., hidden replay, see Section \ref{sec:wakesleep}), there is still concept drift, which can be alleviated using exemplars. Exemplars can be selected in many ways, e.g., via random sampling of the current wake buffer, importance sampling, via clustering of data samples, or via other L2 techniques (e.g.\!, \cite{rebuffi2017icarl}). 
        \item \textbf{Replay Generator}: This is used for training the memory model and skill consolidation by dynamically generating replay samples.
        \item \textbf{Batch Sampler}: The batch sampler usually acts as a random/equal weight sampler, but can have utilize other sampling mechanisms like prioritized replay, which we employ for danger detection as discussed early.
    \end{itemize}

\subsubsection{Student-Teacher Learning}\hfill\\
Various components of the model are trained following a \emph{Student-Teacher Learning} paradigm. The Wake policy model is trained via a RL algorithm of choice. The policy logits are stored in the wake trajectories along with the observations, rewards, actions taken, and other meta-data. The sleep policies are updated (i.e., skill consolidation) using a distillation loss encouraging the sleep model to imitate the (observation, action logit) pairs collected by the Wake model as well as (observation, action logit) pairs sampled from various replay buffers (varies based on system specification). In our experiments, we use experience replay, random exemplar replay, and generative replay. In the case that a generative memory model is used, it is updated via a reconstruction loss comparing raw observations to reconstructed observations or comparing some preprocessed features to reconstructed versions of those features. 
    
Closely tied to the Student-Teacher Learning are \emph{Expert Advice} and the \emph{Skill Selector}. The Expert Advice defines a mixture probability $p_{advice}$ that tells the agent whether to use the current wake policy to process the current observation or whether to use the policy of an expert teacher (generally defined by the sleep model). For the first wake phase, no advice is taken. In subsequent wake phases, the Expert Advice module samples from the sleep model with decaying probability over time. The goal is that this will encourage the wake model to explore in a more intelligent way if there is positive forward transfer between the tasks the sleep model has seen and the current task, ultimately teaching the wake model more effective policies. The Expert Advice probability is set by the Advice Scheduler, which is highly configurable (e.g.\!, constant, linearly decaying, exponentially decaying, cyclic). The time-to-decay is also a configurable parameter. In our experiments, we set it to start with a high probability ($>0.8$) and decay to a low probability (<0.2) by the half-way point of the wake phase's learning period, after which it remains constant until the next wake phase. We have observed this type of scheduler performs well in practice.
    
The sleep model may have multiple sub-policies (e.g., if it consists of a mixture of experts). In this case, a Skill Selector is needed to select which policy should be employed at the start of the wake phase. In order to do this, a small buffer of observations are stored, and tested across multiple sleep policies. The policy which yields the highest reward on this test set is selected to act as the Teacher network for the wake policy. In our experiments, we have also tried a similar approach/variant wherein the weights of the best sleep policy are copied directly to the wake policy for strong initialization of the wake model for the current task, promoting strong forward transfer (as the system is using the current best known policy for the given task, similar to the jump start provided by advice). This is especially true in the case where the model learns distinct policies for individual tasks/skills such as in the case of the mixture of experts or hierarchically-clustered policies.
    


\begin{algorithm}[h]
   \caption{Wake-Sleep Setting}
   \label{alg:wakesleep}
\begin{algorithmic}[1]
   \STATE {\bfseries Iterates:} Generator $g_s^t$, sleep policy $\pi_{s}^t$, wake policy $\pi_{w}^t$, wake buffer $b^t_w$
	   \FOR{$t=1,2,\ldots$}
	   \FOR[Wake Phase]{$K$ times}
	   	   \STATE{Run Skill Selector to select Teacher policy}
	       \STATE{Sample observation $o$ from current task}
	       \STATE{$o_{p} \leftarrow$ Pass observation $o$ through the Preprocessors}
	       \STATE{$o_{a} \leftarrow$ Pass $o_{p}$ through the Annotators}
	       \STATE Using Expert Advice: Sample $a \sim \pi_s^t(o_{a})$ w.p. $p_{\text{advice}}$, else $a \sim \pi_w^t(o_{a})$
	       \STATE{Compute reward $r$ from taking action}
	       \STATE{Add transition $(o_{a}, r, a)$ to Experience Buffer $b_w^t$}
	       \STATE{Update wake policy $\pi^t_w$ on current task reward using $b^t_w$}
		   \STATE{Decay $p_\text{advice}$ according to Advice Scheduler}
	   \ENDFOR\\ 

	   \FOR[Sleep Phase: Skill Consolidation]{$N$ iterations} 
		   \STATE{Sample batch $B$ from $b_w^t$ using Batch Sampler on Experience Replay Buffer.} 
			 \STATE Using Memory Model: Sample batches $O_s^t \sim g_s^t$
			 \STATE Pseudo-label $A_s^t = \pi_s^t(O_s^t)$
			 \STATE (GR Buffer) $B_{S} = B \cup (O_s^t, A_s^t)$
			 \STATE Minimize distillation loss + reconstruction loss on $B_{S}$
		\ENDFOR\\
	 \ENDFOR
\end{algorithmic}
\end{algorithm}

\subsubsection{System Configuration Runner} \hfill\\
The L2RLCF \emph{System Configuration Runner (SCR)} spins off the whole system from a YAML configuration, which contains all of the information needed to instantiate and run the system. 

The SCR has many similarities with hydra \cite{Yadan2019Hydra}, but is specific to our use-case. The SCR enables iteration over multiple experiment settings and quick spin-off of these systems. It has the following abilities:

\begin{itemize}
    \item Can set system parameters through a YAML configuration and also as argparse settings.
    \item Can recursively loop through the YAML configuration to load functors and instantiate class objects, removing the constraint that class arguments are restricted to \emph{Built-In} types, giving the system the ability 
    to hierarchically load objects, and allowing for custom parameters and custom class objects.
    \item Can set system environment variables which might be required by environment simulators or by other components where parameter passing through object initiation is not possible. 
    \item Allows for configuration templating. In L2RLCF, many parameters like learning block size or observation-actions space dimension are relevant to multiple components. The template section of the configuration allows parameters to be set (including assigning functors or instantiating \emph{shared} class objects) and have them available across multiple components.
    \item \newedit{Helps with scheduling of components to specific GPU IDs}
\end{itemize}
Unlike hydra, the SCR doesn't have hierarchical loading of hydra-configuration files, but similar feature might be helpful in further improving configuration management for L2RLCF.

\subsubsection{Containerization}\hfill\\
L2RLCF is containerized with docker with all the dependencies of the main system installed by default. New configurations are mounted to docker for running the experiments. One design decision we took was to go for a single docker (monolith) as opposed to a pod of dockers (microservices). The argument for going with a microservices architecture is that many of the components are independently developed across different institutions with their own  dependencies and  system needs. By maintaining a single docker container, we avoid fractured and inefficient structuring of the system, ensure consistent and standardized versioning of external libraries, avoid decreases in speed resulting from communication through network protocols, and enable end-to-end training across the multiple system components.



\section{Evaluation Environment}
\label{sec:evaluation_env}
While our proposed framework/API is compatible with generic lifelong learning reinforcement learning settings, we introduce an evaluation environment that is sufficient complex while designed for fair, comprehensive, and quantitative comparison of different combinations of components within a common evaluation environment. Our evaluation environment employs different lifelong learning scenarios (sequences of tasks) consisting of \SC minigames derived from \citep{vinyals2017starcraft}. \SC is a real-time strategy game where a player must manage multiple units in combat, collection, and construction tasks to defeat an enemy opponent. In our evaluation environment, the RL agent has control over selecting units and directing the actions the unit should take to accomplish a given task. In the L2RL setting, the system must learn to solve one task at a time without forgetting previous tasks, and the agents performance is measured on all tasks immediately after learning a task. We selected three minigames with two variants each as our task set. Each task involves either i) combat between different unit types or ii) resource collection (\emph{CollectMineralShards}). The tasks include:
\begin{itemize}
    \item \textbf{Collect Mineral Shards – No Fog of War}: A map with 2 Marines and an endless supply of Mineral Shards. Rewards are earned by moving the Marines to collect the Mineral Shards. Whenever all 20 Mineral Shards have been collected, a new set of 20 Mineral Shards are spawned at random locations (at least 2 units away from all Marines). Fog of war is disabled. 
    \item \textbf{Collect Mineral Shards – Fog of war:} A map with 2 Marines and an endless supply of Mineral Shards. Rewards are earned by moving the Marines to collect the Mineral Shards. Whenever all 20 Mineral Shards have been collected, a new set of 20 Mineral Shards are spawned at random locations (at least 2 units away from all Marines). Fog of war is enabled, meaning the agent must be able to learn without full knowledge of the current state of the environment.
    \item \textbf{DefeatZerglingsAndBanelings – One Group:} A map with 9 Marines on the opposite side from a group of 6 Zerglings and 4 Banelings. Rewards are earned by using the Marines to defeat Zerglings and Banelings. Whenever all Zerglings and Banelings have been defeated, a new group of 6 Zerglings and 4 Banelings is spawned, and the player is awarded 4 additional Marines at full health, with all other surviving Marines retaining their existing health (no restore). Whenever new units are spawned, all unit positions are reset to opposite sides of the map.
    \item \textbf{DefeatZerglingsAndBanelings – Two Groups:} A map with 9 Marines in the center with 2 groups consisting of 9 Zerglings on one side and 6 Banelings on the other side. Rewards are earned by using the Marines to defeat Zerglings and Banelings. Whenever a group has been defeated, a new group of 9 Zerglings and 6 Banelings is spawned and the player is awarded 6 additional Marines at full health, with all other surviving Marines retaining their existing health (no restore). Whenever new units are spawned, all unit positions are reset to opposite sides of the map.
    \item \textbf{DefeatRoaches – One Group:} A map with 9 Marines and a group of 4 Roaches on opposite sides. Rewards are earned by using the Marines to defeat Roaches. Whenever all 4 Roaches have been defeated, a new group of 4 Roaches is spawned and the player is awarded 5 additional Marines at full health, with all other surviving Marines retaining their existing health (no restore). Whenever new units are spawned, all unit positions are reset to opposite sides of the map.
    \item \textbf{DefeatRoaches – Two Groups:} A map with 9 Marines in the center and 2 groups consisting of 6 total Roaches on opposite sides (3 on each side). Rewards are earned by using the Marines to defeat Roaches. Whenever all 6 Roaches have been defeated, a new group of 6 Roaches is spawned and the player is awarded 7 additional Marines at full health, with all other surviving Marines retaining their existing health (no restore). Whenever new units are spawned, all unit positions are reset to starting areas of the map.
\end{itemize}

PySC2 \citep{vinyals2017starcraft} was used to interface with SC-2. For the hand-crafted observation space, We used a subset of the available observation maps: the unit type, selection status, and unit density two-dimensional observations. The action space is factored into functions and arguments, such as $\text{move}(x, y)$ or $\text{stop}()$. The agent receives positive rewards for collecting resources and defeating enemy units and negative rewards for losing friendly units. For our experiments, we consider syllabi consisting of alternating (two tasks, each seen three times) and condensed (all six tasks, each seen once) scenarios. 


\subsection{Metrics for Lifelong RL}
To quantitatively evaluate the performance of a L2RL system, we consider two sets of metrics. First, we consider how the rewards achieved by an agent compare to the ``optimal'' RL agent by comparing to the terminal performance of agents trained on each task to convergence (a ``single task expert''). These ``relative reward (RR)'' to the terminal reward achieved by a single task expert metrics are introduced in \cite{daniels2022model}. Note that these metrics focus purely on understanding the dynamics of an agent at periodic evaluation blocks (EBs). Second, we compare algorithms using the lifelong learning metrics defined by New et al. in \cite{new2022lifelong}, which take into account both behavior of the agent at periodic evaluation blocks but also characteristics of the agent as it learns (during learning blocks ``LBs'') (i.e., its learning curves during wake).


We consider the following variants of the RR metric:
\textbf{Relative reward in the final EB (RR\textsubscript{$\Omega$}):} Measures how well the agent performs on all tasks after completing the syllabi. 
\textbf{Relative reward on known tasks (RR\textsubscript{$\sigma$}):} Measures how well the agent performs on previously seen tasks (quantifies forgetting/ backward transfer.
\textbf{Relative reward on unknown tasks (RR\textsubscript{$\upsilon$}):} Measures how well the agent generalizes/transfers knowledge from seen to unseen tasks. Note that in all cases, more-positive values are better for all metrics.


We consider the following lifelong learning metrics defined by \citet{new2022lifelong}: \textbf{Forward Transfer Ratio (FTR):} Measures knowledge transfer to \emph{unknown} tasks.
\textbf{Backward Transfer Ratio (BTR):} Measures knowledge transfer to \emph{known} tasks. A value greater than one indicates positive transfer. 
\textbf{Relative Performance (RP):} Compares the learning curves between the lifelong learner and a single task learner. A value greater than one indicates either faster learning by the lifelong learner and/or superior asymptotic performance. 
\textbf{Performance Maintenance (PM):} Measures catastrophic forgetting over the entire syllabus. A value less than $0$ indicates forgetting.

\begin{table*}[]
    \centering
    \resizebox{0.81\textwidth}{!}{
    \begin{tabular}{|c|c|c|c|c|c|c|c|c|c|}
            \hline
            Scenario & Agent & $PM$ & $FT$ & $BT$ & $RP$ & $RR_{\Omega}$ & $RR_{\sigma}$ & $RR_{\upsilon}$ & \newedit{$Amortized~Phase~Run$-$Time~(s)$} \\
            \hline
            \multirow{8}{*}{\rotatebox[origin=c]{90}{Alternating}}
            & Baseline Vtrace                & $-8.99$ ($ \pm 7.23$) & $1.11$ ($ \pm 0.55$) & $0.79$ ($ \pm 0.30$) & $0.92$ ($ \pm 0.11$) & $0.90$ & $0.82$ & $0.39$ & $12,574$ ($ \pm \; 1,405$) \\
            & Hidden Replay                  & $-6.13$ ($ \pm 7.31$)  & $1.85$ ($ \pm 1.38$)  & $0.87$ ($ \pm 0.20$)  & $0.91$ ($ \pm 0.13$) & $0.82$ & $0.78$ & $0.57$ & $32,297$ ($ \pm \; 3,646$) \\
            & Hidden Replay + REMIND         & $-0.56$ ($ \pm 0.79$)  & $0.96$ ($ \pm 0.53$)  & $0.95$ ($ \pm 0.15$)  & $0.58$ ($ \pm 0.19$) & $0.53$ & $0.51$ & $0.29$ & $50,766$ ($ \pm \; 5,037$)\\
            & Hidden Replay + Adaptive Sleep & $-5.00$ ($ \pm 3.69$)  & $1.11$ ($ \pm 0.38$)  & $0.87$ ($ \pm 0.11$)  & $0.84$ ($ \pm 0.13$) & $0.78$ & $0.75$ & $0.39$ & $29,932$ ($ \pm \; 5,325$)\\
            & Hidden Replay + SSRL           & $-8.05$ ($ \pm 6.22$)  & $1.16$ ($ \pm 0.62$)  & $0.79$ ($ \pm 0.13$)  & $0.97$ ($ \pm 0.21$) & $0.87$ & $0.92$ & $0.43$ & $52,346$ ($ \pm 10,542$)\\
            \cline{2-10}
            & Baseline Vtrace (Danger Tasks) & $-1.32$ ($ \pm 1.62$) & $1.62$ ($ \pm 0.46$) & $0.96$ ($ \pm 0.08$) & $1.02$ ($ \pm 0.03$) & $1.13$ & $1.06$ & $0.76$ & $13,349$ ($ \pm \;\;\; 766$) \\
            & Hidden Replay (Danger Tasks)   & $-0.65$ ($ \pm 1.51$)  & $2.45$ ($ \pm 1.78$)  & $1.01$ ($ \pm 0.07$)  & $1.01$ ($ \pm 0.08$) & $1.00$ & $0.96$ & $0.80$ & $33,307$ ($ \pm \; 3,270$) \\
            & Hidden Replay + Danger Det.    & $-0.78$ ($ \pm 1.57$)  & $1.64$ ($ \pm 0.79$)  & $0.98$ ($ \pm 0.10$)  & $0.97$ ($ \pm 0.10$) & $1.05$ & $0.97$ & $0.67$ & $54,088$ ($ \pm \; 3,992$)\\
            \hline
            \multirow{7}{*}{\rotatebox[origin=c]{90}{Condensed}}
            & Baseline Vtrace                & $-3.41$ ($ \pm 4.03$) & $1.19$ ($ \pm 0.22$) & $1.17$ ($ \pm 0.69$) & $1.14$ ($ \pm 0.14$) & $0.64$ & $0.66$ & $0.49$ & $12,491$ ($ \pm \; 1,389$) \\
            & Hidden Replay                  & $-3.05$ ($ \pm 1.76$)  & $1.42$ ($ \pm 0.11$)  & $1.00$ ($ \pm 0.03$)  & $1.17$ ($ \pm 0.11$) & $0.80$ & $0.83$ & $0.60$ & $32,284$ ($ \pm \; 3,954$) \\
            & Hidden Replay + REMIND         & $-0.15$ ($ \pm 1.54$)  & $1.17$ ($ \pm 0.06$)  & $1.02$ ($ \pm 0.07$)  & $0.77$ ($ \pm 0.05$) & $0.56$ & $0.55$ & $0.45$ & $41,688$ ($ \pm 12,281$)\\
            & Hidden Replay + Adaptive Sleep & $-2.23$ ($ \pm 2.37$)  & $1.26$ ($ \pm 0.09$)  & $1.04$ ($ \pm 0.11$)  & $1.08$ ($ \pm 0.10$) & $0.72$ & $0.74$ & $0.53$ & $28,545$ ($ \pm \; 6,249$)\\
            & Hidden Replay + SSRL           & $-5.91$ ($ \pm 4.05$)  & $1.32$ ($ \pm 0.15$)  & $1.10$ ($ \pm 0.13$)  & $1.13$ ($ \pm 0.14$) & $0.76$ & $0.79$ & $0.48$ & $60,282$ ($ \pm 13,276$)\\
            \cline{2-10}
            & Hidden Replay (Danger Tasks)   & $-0.63$ ($ \pm 0.77$) & $1.44$ ($ \pm 0.27$) & $0.99$ ($ \pm 0.04$) & $1.08$ ($ \pm 0.09$) & $0.95$ & $0.99$ & $0.80$ & $24,921$ ($ \pm \; 2,762$) \\
            & Hidden Replay + Danger Det.    & $-0.86$ ($ \pm 1.92$)  & $1.45$ ($ \pm 0.25$)  & $0.97$ ($ \pm 0.06$)  & $1.09$ ($ \pm 0.08$) & $0.97$ & $1.01$ & $0.77$ & $52,751$ ($ \pm \; 6,265$)\\
            \hline
    \end{tabular}
    }
    \caption{Lifelong RL metrics of integrated case study agents. All experiments standardized with base architecture and 2 million RL steps per task
    }
    \label{tab:extensions_results}
\end{table*}
\section{Case Study: Integration of Multiple Lifelong Learning Algorithms in a Unified System}
\label{sec:case_study}
In this section, we discuss the integration of multiple lifelong learning algorithms using our L2RLCF framework and API in a fully-realized real-world system. In this case study, we integrate the following algorithmic components:
\begin{itemize}
    \item Base system for L2RL based on generative replay in a wake-sleep setting
    \item Automatic triggering of sleep via changepoint detection
    \item Compression of experiences in the replay buffer
    \item Task-specific prioritized replay mechanism for dangerous state detection
    \item Representation learning from RGB-observations via self-supervised learning
\end{itemize}
\textbf{Note that our system is not limited to these components. It is flexible to be used with any lifelong learning algorithm that can be integrated within a wake-sleep mechanism.} To provide context for the complexity of the integration effort and flexibility of L2RLCF, the details of each component used in this case study is described in the following sections.

\subsection{Standalone Optimal policy Trajectory Dataset}
\newedit{We learn single task experts (STEs) by running the RL policy in just the wake phase for a given task. Since deep RL in \SC is computationally demanding, we release a dataset of trajectories of trained single-task expert policies for future research \footnote{\url{https://github.com/sri-l2m/l2m\_data}}. We curate these trajectories to create the Optimal Policy Trajectory dataset. This dataset is used for the standalone pretraining of some of the components that are discussed in this section.}

\subsection{Wake-Sleep Generative Replay Algorithm}
\label{sec:wakesleep}

As previously mentioned, the fundamental tradeoff in lifelong learning is between \emph{plasticity}, i.e.\! learning the current task, and \emph{stability}, i.e.\! remembering past tasks. To address this tradeoff in the context of lifelong RL, we extend the wake-sleep mechanism first introduced by Raghavan et al.\! in \cite{raghavan2020lifelong}.
This approach utilizes two phases:
\begin{itemize}
    \item \textbf{Wake Phase:} A \emph{plastic} wake policy $\pi_w$ is optimized for the current task by interacting with the environment and using an off-the-shelf RL method. Transition tuples are collected during training and stored in a buffer; each tuple contains $(o, r, a)$, the observation $o$, reward for the previous action $r$, and the policy output $a$ (e.g., the policy logits or one-hot encoded action). The sleep policy $\pi_s$ provides ``advice'' (with importance decaying over time) in order to encourage the wake RL agent to begin exploring the current task using the consolidated policy learned from all previous tasks to encourage faster adaptation to the new task. In our experiments, we use an off-policy RL algorithm such as VTrace \cite{espeholt2018impala} to accommodate this off-policy action selection in the optimization of $\pi_w$.
    \item \textbf{Sleep Phase:} In the sleep phase, the \emph{stable} sleep policy $\pi_s$ is optimized to maximize the incorporation of new knowledge (the action selection in the wake buffer) while minimally forgetting current knowledge. While not a general requirement of the wake-sleep mechanism, in the particular implementation of wake-sleep used in this study, we also employ an additional replay type akin to generative replay in supervised learning. The augmented dataset(s) are created by combining wake transitions with tuples from a generative model $g_s$, which generates observations that are subsequently pseudo-labelled by the previous sleep policy. The sleep policy and the generative model are jointly trained.
\end{itemize}
Our base model has a unique architecture. In contrast to most generative replay-based models which reconstruct the observations in observation space, the architecture we use for testing purposes consists of a model that learns separate branches for i) reconstructing/generating intermediate hidden states from a feature extractor and ii) a policy network for predicting which action to take given an observation (while sharing a common feature extractor). We call this the ``Hidden Replay Architecture''. More details of this base model can be found in \cite{daniels2022model}. It should be noted that our system can work with any architecture/replay mechanism as long as it is implemented in a wake-sleep setting, and in fact, we have validated our integrated system with other, more traditional architectures, but do not report results here.



\subsection{Self-Monitoring for Sleep via WATCH}
\label{sec:self-monitoring}
In the base model, sleep is triggered after a fixed number of interactions with the environment. This can be inefficient due to the overhead of sleep, and lead to reduced performance due to over emphasizing memory consolidation versus learning. We hypothesize that triggering sleep adaptively at opportune times will lead to better performance with less overhead. We apply an unsupervised change-point detection method \cite{faber2021watch, faber2022lifewatch} to the features extracted from SC-2 observations using a pre-trained VGG-16 model \cite{simonyan2014very}
because (a) our setting does not assume knowledge of task change points, and (b) sleep can be beneficial even when the task has not changed, i.e. a significant change in the policy can cause it to visit novel states and observations. In principle, change point detection can be applied to episodic reward as well.  

Preliminary experiments showed that standard methods for change point detection like CUSUM appear unreliable in the presence of  the high-dimensional features, whereas the recently proposed (LIFE)WATCH \cite{faber2021watch,faber2022lifewatch} method performs better. It has a few crucial benefits exploited in the system. First, it compares two sets of points instead of assuming any specific distribution, providing a more flexible approach. Second, leveraging Wasserstein distance allows for more accurate and robust detection in dynamic high-dimensional data. Moreover, instead of applying the absolute constants threshold, WATCH adapts to the discovered distribution learning it over time.
\newedit{This component is added as preprocessor block into the framework. The changepoint detector checks for changes in the observation space and signals the system to go to sleep if needed.}

\subsection{Experience Compression via REMIND} 
Clever strategies for efficient management of replay buffers has been shown to improve supervised continual learning \cite{riemer2019scalable, wang2021acae} and lifelong RL \cite{caccia2020online,zhang2017deeper,schaul2015prioritized}.
In contrast to prior work that rejects most transitions from being stored in the buffer, we explore an approach that stores all the transitions but in a highly compressed form

Compression allows a buffer of a given size to contain a larger number of transitions, increasing the chance of retaining diverse examples from different tasks. We use the REMIND \cite{hayes2019remind} method, based on Product Quantization (PQ) \cite{jegou2010product}, to compress the agent's observations before storing them in the replay buffer, allowing the system to store more samples. The agent observations are image-like \texttt{float32} tensors with $3$ channels, and thus each pixel occupies $96$ bits. In the compressed observation, each pixel is quantized to an 8-bit integer, a 24x reduction in size. The PQ model is pre-trained on observations collected while following a random policy on a subset of tasks that is disjoint from the set of evaluation tasks. 
\newedit{This component is also integrated as a preprocessor. It quantizes the current observation and makes it available for the downstream task.}

\subsection{Danger Detection for Prioritized Replay} 
\label{sec:danger}
In addition to compression of experiences, we examine a novel form of prioritized experience replay \cite{schaul2015prioritized} based on detecting dead-end states (expressed as dangerous states in our \SC evaluation environment). We hypothesize that increasing the lifetime of an agent by avoiding dead end states is a useful bias. The Danger Detector outputs a ``danger score'' of how likely the agent is to lose the battle from a given state. This score is used as a replay priority. The policy's actions in the battle change over time, making the danger detection task a continual learning problem. We used Deep Streaming Linear Discriminant Analysis (DeepSLDA) \cite{hayes2019lifelong} for our danger detector. At the end of each episode, we obtain ground truth of the result of the battle and continually update the danger detector. 
Deep SLDA works on top of a fixed feature extractor; we pre-trained a feature extractor based on the FullyConv architecture of \citet{vinyals2017starcraft} using data generated from single task experts (agents trained to convergence using a standard RL algorithm for a single task).
\newedit{This component is integrated as an annotator block. The danger detector annotates the observations on the likelihood of if the state is dangerous; by following safe policies during wake and biasing the data collection process, this amounts to a form of prioritized replay used during the sleep phase's memory consolidation.}


\subsection{Self-Supervised Representation Learning for Generalization}

The \SC evaluation framework can represent observations either as hand-crafted representations provided by the PySC2 simulator or as RGB depictions of the screen. Our proposed system is capable of operating over both types of representation. By default, the system operates over the first type of representation. In practice, agents must learn a  representation from RGB percepts as tasks change over time. In this section, we consider self-supervised continual representation learning. 

Motivated by recent work that representation learned using self-supervision auxiliary losses can boost lifelong learning performance~\cite{gallardo2021self}. The intuition is that features learned from an auxiliary losses generalize more than features learned from task-specific losses, and therefore may be less vulnerable to forgetting. The representation is generated from RGB input from two SC-2 tasks, processed by a ResNet18~\cite{he2016deep}, then refined using Barlow Twins~\cite{zbontar2021barlow} self-supervised learning. We chose this approach by validating on object detection performance using Average Precision 50(AP50) as a metric within the SC-2 framework, finding that 1) the ImageNet1k pre-training was beneficial and 2) Barlow Twins outperformed other SSL methods such as MoCo-V2~\cite{chen2020improved} and SimCLR~\cite{chen2020simple}. 
\newedit{This component is also integrated as a preprocessor where the self-supervised features of the observations are extracted and made available for the downstream tasks.}

\section{Experiments}

\subsection{Integrated SC-2 Agents}

In this section, we demonstrate how the evaluation environment is useful for understanding the effects of different components. We set up an experimental environment for evaluating the components discussed in Section \ref{sec:case_study} using standardized sets of syllabi for different scenarios under identical wake/sleep conditions with the same base agent. We show results in Table \ref{tab:extensions_results} where we specifically look at the case of turning on one additional component at a time. This enables the user of the lifelong learning system to understand the advantages and disadvantages of adding one or more components. For example, we can see that compressing the replay can negatively affect the performance of the system in terms of rewards compared to the single task expert in most cases, and oppositely, self-supervised learning generally helps w.r.t. performance relative to the single task expert. Similarly, we can see adaptive sleep can often improve performance maintenance.

\subsection{Agent Run-Time}

All the experiments are run with the wake actors interacting with 8 \SC simulators. Each wake phase has 2 million interaction steps with 2 forced sleep phase in between. For experiments involving adaptive sleep, the number of sleep phases in a given learning block is dynamic. 

Note that in this work, we don't target developing the most efficient system. Instead, our goal is to develop a framework for easily combining lifelong learning algorithms and an evaluation setting for understanding the trade-offs associated with each component/combination of components. One important trade-off is understanding how adding a component improves the performance of the system at the cost of increased run-time.

The amortized phase run-times, in seconds, are reported in Table \ref{tab:extensions_results} along with performance metrics.
As seen from the table, adding different components increases the run-time for each phase, but the learners themselves become more sample efficient, i.e., requiring about 2 million steps to learn a given task as opposed to about 10 million steps needed for single task experts. 


\section{Discussion}
In this work, we introduced a common framework for integrating diverse lifelong learning components in a unified system. The system has just one assumption, i.e. only requiring the use of a wake-sleep cycle. To deploy the system in the real world, we developed a well-defined API. We also introduced a challenging evaluation environment to fairly assess the impact of integrating multiple components and assessing their effect on the overall system in a quantitative way. To demonstrate that the framework is useful in practical scenarios, we constructed a case study integrating multiple complex existing lifelong learning algorithms. The experiments showed how we could use the evaluation environment to identify some of the strengths and weaknesses of included algorithms. This process can be easily reproduced and allows for the inclusion of new algorithms, providing an effective tool for their analysis. While there is much work to be done to improve the system and ultimately promote its adaptation in academia and industry, we expect that such a system is incredible useful for translating L2RL from research to real-world applications. If adopted, our system could majorly impact some of the domains mentioned earlier: autonomous vehicles, service robots, medicine, and network security among many others, and it could be a useful tool for minimizing model obsolescence and promoting fast model adaptation in dynamically-changing environments.

\begin{acks}
  This material is based upon work supported by the Defense Advanced Research Projects Agency (DARPA) under the Lifelong Learning Machines (L2M) program Contract No. HR0011-18-C-0051. Any opinions, findings and conclusions or recommendations expressed in this material are those of the authors and do not necessarily reflect the views of the Defense Advanced Research Projects Agency (DARPA).

  Special thanks to Constantine Dovrolis for his useful discussions and feedback related to system design.
\end{acks}
\newline

\bibliographystyle{ACM-Reference-Format}
\bibliography{main}


\begin{thebibliography}{39}


\ifx \showCODEN    \undefined \def \showCODEN     #1{\unskip}     \fi
\ifx \showDOI      \undefined \def \showDOI       #1{#1}\fi
\ifx \showISBNx    \undefined \def \showISBNx     #1{\unskip}     \fi
\ifx \showISBNxiii \undefined \def \showISBNxiii  #1{\unskip}     \fi
\ifx \showISSN     \undefined \def \showISSN      #1{\unskip}     \fi
\ifx \showLCCN     \undefined \def \showLCCN      #1{\unskip}     \fi
\ifx \shownote     \undefined \def \shownote      #1{#1}          \fi
\ifx \showarticletitle \undefined \def \showarticletitle #1{#1}   \fi
\ifx \showURL      \undefined \def \showURL       {\relax}        \fi
\providecommand\bibfield[2]{#2}
\providecommand\bibinfo[2]{#2}
\providecommand\natexlab[1]{#1}
\providecommand\showeprint[2][]{arXiv:#2}

\bibitem[\protect\citeauthoryear{??}{wik}{2022}]%
        {wikipedia_2022}
 \bibinfo{year}{2022}\natexlab{}.
\newblock \bibinfo{title}{Companion Robot}.
\newblock
\newblock
\urldef\tempurl%
\url{https://en.wikipedia.org/wiki/Companion_robot}
\showURL{%
\tempurl}


\bibitem[\protect\citeauthoryear{Alahdab and {\c{C}}al{\i}kl{\i}}{Alahdab and
  {\c{C}}al{\i}kl{\i}}{2019}]%
        {alahdab2019empirical}
\bibfield{author}{\bibinfo{person}{Mohannad Alahdab} {and}
  \bibinfo{person}{G{\"u}l {\c{C}}al{\i}kl{\i}}.}
  \bibinfo{year}{2019}\natexlab{}.
\newblock \showarticletitle{Empirical Analysis of Hidden Technical Debt
  Patterns in Machine Learning Software}. In
  \bibinfo{booktitle}{\emph{International Conference on Product-Focused
  Software Process Improvement}}. Springer, \bibinfo{pages}{195--202}.
\newblock


\bibitem[\protect\citeauthoryear{Bogner, Verdecchia, and
  Gerostathopoulos}{Bogner et~al\mbox{.}}{2021}]%
        {bogner2021characterizing}
\bibfield{author}{\bibinfo{person}{Justus Bogner}, \bibinfo{person}{Roberto
  Verdecchia}, {and} \bibinfo{person}{Ilias Gerostathopoulos}.}
  \bibinfo{year}{2021}\natexlab{}.
\newblock \showarticletitle{Characterizing technical debt and antipatterns in
  ai-based systems: A systematic mapping study}. In
  \bibinfo{booktitle}{\emph{2021 IEEE/ACM International Conference on Technical
  Debt (TechDebt)}}. IEEE, \bibinfo{pages}{64--73}.
\newblock


\bibitem[\protect\citeauthoryear{Caccia, Belilovsky, Caccia, and Pineau}{Caccia
  et~al\mbox{.}}{2020}]%
        {caccia2020online}
\bibfield{author}{\bibinfo{person}{Lucas Caccia}, \bibinfo{person}{Eugene
  Belilovsky}, \bibinfo{person}{Massimo Caccia}, {and} \bibinfo{person}{Joelle
  Pineau}.} \bibinfo{year}{2020}\natexlab{}.
\newblock \showarticletitle{Online learned continual compression with adaptive
  quantization modules}. In \bibinfo{booktitle}{\emph{International Conference
  on Machine Learning}}. PMLR, \bibinfo{pages}{1240--1250}.
\newblock


\bibitem[\protect\citeauthoryear{Chen, Kornblith, Norouzi, and Hinton}{Chen
  et~al\mbox{.}}{2020b}]%
        {chen2020simple}
\bibfield{author}{\bibinfo{person}{Ting Chen}, \bibinfo{person}{Simon
  Kornblith}, \bibinfo{person}{Mohammad Norouzi}, {and}
  \bibinfo{person}{Geoffrey Hinton}.} \bibinfo{year}{2020}\natexlab{b}.
\newblock \showarticletitle{A simple framework for contrastive learning of
  visual representations}. In \bibinfo{booktitle}{\emph{International
  conference on machine learning}}. PMLR, \bibinfo{pages}{1597--1607}.
\newblock


\bibitem[\protect\citeauthoryear{Chen, Fan, Girshick, and He}{Chen
  et~al\mbox{.}}{2020a}]%
        {chen2020improved}
\bibfield{author}{\bibinfo{person}{Xinlei Chen}, \bibinfo{person}{Haoqi Fan},
  \bibinfo{person}{Ross Girshick}, {and} \bibinfo{person}{Kaiming He}.}
  \bibinfo{year}{2020}\natexlab{a}.
\newblock \showarticletitle{Improved baselines with momentum contrastive
  learning}.
\newblock \bibinfo{journal}{\emph{arXiv preprint arXiv:2003.04297}}
  (\bibinfo{year}{2020}).
\newblock


\bibitem[\protect\citeauthoryear{Daniels, Raghavan, Hostetler, Rahman, Sur,
  Piacentino, and Divakaran}{Daniels et~al\mbox{.}}{2022}]%
        {daniels2022model}
\bibfield{author}{\bibinfo{person}{Zachary Daniels}, \bibinfo{person}{Aswin
  Raghavan}, \bibinfo{person}{Jesse Hostetler}, \bibinfo{person}{Abrar Rahman},
  \bibinfo{person}{Indranil Sur}, \bibinfo{person}{Michael Piacentino}, {and}
  \bibinfo{person}{Ajay Divakaran}.} \bibinfo{year}{2022}\natexlab{}.
\newblock \showarticletitle{Model-Free Generative Replay for Lifelong
  Reinforcement Learning: Application to Starcraft-2}. In
  \bibinfo{booktitle}{\emph{Conference on Lifelong Learning Agents}}.
  Proceedings of Machine Learning Research.
\newblock


\bibitem[\protect\citeauthoryear{De~Lange, Aljundi, Masana, Parisot, Jia,
  Leonardis, Slabaugh, and Tuytelaars}{De~Lange et~al\mbox{.}}{2021}]%
        {de2021continual}
\bibfield{author}{\bibinfo{person}{Matthias De~Lange}, \bibinfo{person}{Rahaf
  Aljundi}, \bibinfo{person}{Marc Masana}, \bibinfo{person}{Sarah Parisot},
  \bibinfo{person}{Xu Jia}, \bibinfo{person}{Ale{\v{s}} Leonardis},
  \bibinfo{person}{Gregory Slabaugh}, {and} \bibinfo{person}{Tinne
  Tuytelaars}.} \bibinfo{year}{2021}\natexlab{}.
\newblock \showarticletitle{A continual learning survey: Defying forgetting in
  classification tasks}.
\newblock \bibinfo{journal}{\emph{IEEE transactions on pattern analysis and
  machine intelligence}} \bibinfo{volume}{44}, \bibinfo{number}{7}
  (\bibinfo{year}{2021}), \bibinfo{pages}{3366--3385}.
\newblock


\bibitem[\protect\citeauthoryear{Espeholt, Soyer, Munos, Simonyan, Mnih, Ward,
  Doron, Firoiu, Harley, Dunning, et~al\mbox{.}}{Espeholt
  et~al\mbox{.}}{2018}]%
        {espeholt2018impala}
\bibfield{author}{\bibinfo{person}{Lasse Espeholt}, \bibinfo{person}{Hubert
  Soyer}, \bibinfo{person}{Remi Munos}, \bibinfo{person}{Karen Simonyan},
  \bibinfo{person}{Vlad Mnih}, \bibinfo{person}{Tom Ward},
  \bibinfo{person}{Yotam Doron}, \bibinfo{person}{Vlad Firoiu},
  \bibinfo{person}{Tim Harley}, \bibinfo{person}{Iain Dunning},
  {et~al\mbox{.}}} \bibinfo{year}{2018}\natexlab{}.
\newblock \showarticletitle{Impala: Scalable distributed deep-rl with
  importance weighted actor-learner architectures}. In
  \bibinfo{booktitle}{\emph{International Conference on Machine Learning}}.
  PMLR, \bibinfo{pages}{1407--1416}.
\newblock


\bibitem[\protect\citeauthoryear{Faber, Corizzo, Sniezynski, Baron, and
  Japkowicz}{Faber et~al\mbox{.}}{2021}]%
        {faber2021watch}
\bibfield{author}{\bibinfo{person}{Kamil Faber}, \bibinfo{person}{Roberto
  Corizzo}, \bibinfo{person}{Bartlomiej Sniezynski}, \bibinfo{person}{Michael
  Baron}, {and} \bibinfo{person}{Nathalie Japkowicz}.}
  \bibinfo{year}{2021}\natexlab{}.
\newblock \showarticletitle{WATCH: Wasserstein Change Point Detection for
  High-Dimensional Time Series Data}. In \bibinfo{booktitle}{\emph{2021 IEEE
  International Conference on Big Data (Big Data)}}. IEEE,
  \bibinfo{pages}{4450--4459}.
\newblock


\bibitem[\protect\citeauthoryear{Faber, Corizzo, Sniezynski, Baron, and
  Japkowicz}{Faber et~al\mbox{.}}{2022}]%
        {faber2022lifewatch}
\bibfield{author}{\bibinfo{person}{Kamil Faber}, \bibinfo{person}{Roberto
  Corizzo}, \bibinfo{person}{Bartlomiej Sniezynski}, \bibinfo{person}{Michael
  Baron}, {and} \bibinfo{person}{Nathalie Japkowicz}.}
  \bibinfo{year}{2022}\natexlab{}.
\newblock \showarticletitle{LIFEWATCH: Lifelong Wasserstein Change Point
  Detection}. In \bibinfo{booktitle}{\emph{2022 International Joint Conference
  on Neural Networks (IJCNN)}}. IEEE.
\newblock


\bibitem[\protect\citeauthoryear{Fendley, Costello, Nguyen, Perrotta, and
  Lowman}{Fendley et~al\mbox{.}}{2022}]%
        {fendley2022continual}
\bibfield{author}{\bibinfo{person}{Neil Fendley}, \bibinfo{person}{Cash
  Costello}, \bibinfo{person}{Eric Nguyen}, \bibinfo{person}{Gino Perrotta},
  {and} \bibinfo{person}{Corey Lowman}.} \bibinfo{year}{2022}\natexlab{}.
\newblock \showarticletitle{Continual Reinforcement Learning with TELLA}. In
  \bibinfo{booktitle}{\emph{Workshop on Lifelong Learning Agents}}.
\newblock


\bibitem[\protect\citeauthoryear{Gallardo, Hayes, and Kanan}{Gallardo
  et~al\mbox{.}}{2021}]%
        {gallardo2021self}
\bibfield{author}{\bibinfo{person}{Jhair Gallardo}, \bibinfo{person}{Tyler~L
  Hayes}, {and} \bibinfo{person}{Christopher Kanan}.}
  \bibinfo{year}{2021}\natexlab{}.
\newblock \showarticletitle{Self-supervised training enhances online continual
  learning}.
\newblock \bibinfo{journal}{\emph{arXiv preprint arXiv:2103.14010}}
  (\bibinfo{year}{2021}).
\newblock


\bibitem[\protect\citeauthoryear{Hayes, Kafle, Shrestha, Acharya, and
  Kanan}{Hayes et~al\mbox{.}}{2020}]%
        {hayes2019remind}
\bibfield{author}{\bibinfo{person}{Tyler~L Hayes}, \bibinfo{person}{Kushal
  Kafle}, \bibinfo{person}{Robik Shrestha}, \bibinfo{person}{Manoj Acharya},
  {and} \bibinfo{person}{Christopher Kanan}.} \bibinfo{year}{2020}\natexlab{}.
\newblock \showarticletitle{REMIND Your Neural Network to Prevent Catastrophic
  Forgetting}. In \bibinfo{booktitle}{\emph{ECCV}}.
\newblock


\bibitem[\protect\citeauthoryear{Hayes and Kanan}{Hayes and Kanan}{2020}]%
        {hayes2019lifelong}
\bibfield{author}{\bibinfo{person}{Tyler~L Hayes} {and}
  \bibinfo{person}{Christopher Kanan}.} \bibinfo{year}{2020}\natexlab{}.
\newblock \showarticletitle{Lifelong Machine Learning with Deep Streaming
  Linear Discriminant Analysis}. In \bibinfo{booktitle}{\emph{CVPR-W}}.
\newblock


\bibitem[\protect\citeauthoryear{He, Zhang, Ren, and Sun}{He
  et~al\mbox{.}}{2016}]%
        {he2016deep}
\bibfield{author}{\bibinfo{person}{Kaiming He}, \bibinfo{person}{Xiangyu
  Zhang}, \bibinfo{person}{Shaoqing Ren}, {and} \bibinfo{person}{Jian Sun}.}
  \bibinfo{year}{2016}\natexlab{}.
\newblock \showarticletitle{Deep residual learning for image recognition}. In
  \bibinfo{booktitle}{\emph{Proceedings of the IEEE conference on computer
  vision and pattern recognition}}. \bibinfo{pages}{770--778}.
\newblock


\bibitem[\protect\citeauthoryear{Jegou, Douze, and Schmid}{Jegou
  et~al\mbox{.}}{2010}]%
        {jegou2010product}
\bibfield{author}{\bibinfo{person}{Herve Jegou}, \bibinfo{person}{Matthijs
  Douze}, {and} \bibinfo{person}{Cordelia Schmid}.}
  \bibinfo{year}{2010}\natexlab{}.
\newblock \showarticletitle{Product quantization for nearest neighbor search}.
\newblock \bibinfo{journal}{\emph{TPAMI}}  \bibinfo{volume}{33}
  (\bibinfo{year}{2010}).
\newblock


\bibitem[\protect\citeauthoryear{Johnson, Nguyen, Schreurs, Ewulum, Ashcraft,
  Fendley, Baker, New, and Vallabha}{Johnson et~al\mbox{.}}{2022}]%
        {johnson2022l2explorer}
\bibfield{author}{\bibinfo{person}{Erik~C Johnson}, \bibinfo{person}{Eric~Q
  Nguyen}, \bibinfo{person}{Blake Schreurs}, \bibinfo{person}{Chigozie~S
  Ewulum}, \bibinfo{person}{Chace Ashcraft}, \bibinfo{person}{Neil~M Fendley},
  \bibinfo{person}{Megan~M Baker}, \bibinfo{person}{Alexander New}, {and}
  \bibinfo{person}{Gautam~K Vallabha}.} \bibinfo{year}{2022}\natexlab{}.
\newblock \showarticletitle{L2Explorer: A Lifelong Reinforcement Learning
  Assessment Environment}.
\newblock \bibinfo{journal}{\emph{arXiv preprint arXiv:2203.07454}}
  (\bibinfo{year}{2022}).
\newblock


\bibitem[\protect\citeauthoryear{Khetarpal, Riemer, Rish, and Precup}{Khetarpal
  et~al\mbox{.}}{2020}]%
        {khetarpal2020towards}
\bibfield{author}{\bibinfo{person}{Khimya Khetarpal}, \bibinfo{person}{Matthew
  Riemer}, \bibinfo{person}{Irina Rish}, {and} \bibinfo{person}{Doina Precup}.}
  \bibinfo{year}{2020}\natexlab{}.
\newblock \showarticletitle{Towards continual reinforcement learning: A review
  and perspectives}.
\newblock \bibinfo{journal}{\emph{arXiv preprint arXiv:2012.13490}}
  (\bibinfo{year}{2020}).
\newblock


\bibitem[\protect\citeauthoryear{Kirkpatrick, Pascanu, Rabinowitz, Veness,
  Desjardins, Rusu, Milan, Quan, Ramalho, Grabska-Barwinska, Hassabis, Clopath,
  Kumaran, and Hadsell}{Kirkpatrick et~al\mbox{.}}{2016}]%
        {kirkpatrick_overcoming_2016}
\bibfield{author}{\bibinfo{person}{James Kirkpatrick}, \bibinfo{person}{Razvan
  Pascanu}, \bibinfo{person}{Neil Rabinowitz}, \bibinfo{person}{Joel Veness},
  \bibinfo{person}{Guillaume Desjardins}, \bibinfo{person}{Andrei~A. Rusu},
  \bibinfo{person}{Kieran Milan}, \bibinfo{person}{John Quan},
  \bibinfo{person}{Tiago Ramalho}, \bibinfo{person}{Agnieszka
  Grabska-Barwinska}, \bibinfo{person}{Demis Hassabis},
  \bibinfo{person}{Claudia Clopath}, \bibinfo{person}{Dharshan Kumaran}, {and}
  \bibinfo{person}{Raia Hadsell}.} \bibinfo{year}{2016}\natexlab{}.
\newblock \showarticletitle{Overcoming catastrophic forgetting in neural
  networks}.
\newblock \bibinfo{journal}{\emph{arXiv:1612.00796 [cs, stat]}}
  (\bibinfo{year}{2016}).
\newblock
\urldef\tempurl%
\url{arxiv.org/abs/1612.00796}
\showURL{%
\tempurl}


\bibitem[\protect\citeauthoryear{Ma, Wang, Yang, and Yang}{Ma
  et~al\mbox{.}}{2020}]%
        {ma2020artificial}
\bibfield{author}{\bibinfo{person}{Yifang Ma}, \bibinfo{person}{Zhenyu Wang},
  \bibinfo{person}{Hong Yang}, {and} \bibinfo{person}{Lin Yang}.}
  \bibinfo{year}{2020}\natexlab{}.
\newblock \showarticletitle{Artificial intelligence applications in the
  development of autonomous vehicles: a survey}.
\newblock \bibinfo{journal}{\emph{IEEE/CAA Journal of Automatica Sinica}}
  \bibinfo{volume}{7}, \bibinfo{number}{2} (\bibinfo{year}{2020}),
  \bibinfo{pages}{315--329}.
\newblock


\bibitem[\protect\citeauthoryear{Mai, Li, Jeong, Quispe, Kim, and Sanner}{Mai
  et~al\mbox{.}}{2022}]%
        {mai2022online}
\bibfield{author}{\bibinfo{person}{Zheda Mai}, \bibinfo{person}{Ruiwen Li},
  \bibinfo{person}{Jihwan Jeong}, \bibinfo{person}{David Quispe},
  \bibinfo{person}{Hyunwoo Kim}, {and} \bibinfo{person}{Scott Sanner}.}
  \bibinfo{year}{2022}\natexlab{}.
\newblock \showarticletitle{Online continual learning in image classification:
  An empirical survey}.
\newblock \bibinfo{journal}{\emph{Neurocomputing}}  \bibinfo{volume}{469}
  (\bibinfo{year}{2022}), \bibinfo{pages}{28--51}.
\newblock


\bibitem[\protect\citeauthoryear{Mendez, Wang, and Eaton}{Mendez
  et~al\mbox{.}}{2020}]%
        {mendez2020lifelong}
\bibfield{author}{\bibinfo{person}{Jorge Mendez}, \bibinfo{person}{Boyu Wang},
  {and} \bibinfo{person}{Eric Eaton}.} \bibinfo{year}{2020}\natexlab{}.
\newblock \showarticletitle{Lifelong policy gradient learning of factored
  policies for faster training without forgetting}.
\newblock \bibinfo{journal}{\emph{Advances in Neural Information Processing
  Systems}}  \bibinfo{volume}{33} (\bibinfo{year}{2020}),
  \bibinfo{pages}{14398--14409}.
\newblock


\bibitem[\protect\citeauthoryear{New, Baker, Nguyen, and Vallabha}{New
  et~al\mbox{.}}{2022}]%
        {new2022lifelong}
\bibfield{author}{\bibinfo{person}{Alexander New}, \bibinfo{person}{Megan
  Baker}, \bibinfo{person}{Eric Nguyen}, {and} \bibinfo{person}{Gautam
  Vallabha}.} \bibinfo{year}{2022}\natexlab{}.
\newblock \showarticletitle{Lifelong Learning Metrics}.
\newblock \bibinfo{journal}{\emph{arXiv preprint arXiv:2201.08278}}
  (\bibinfo{year}{2022}).
\newblock


\bibitem[\protect\citeauthoryear{Padakandla}{Padakandla}{2021}]%
        {padakandla2021survey}
\bibfield{author}{\bibinfo{person}{Sindhu Padakandla}.}
  \bibinfo{year}{2021}\natexlab{}.
\newblock \showarticletitle{A survey of reinforcement learning algorithms for
  dynamically varying environments}.
\newblock \bibinfo{journal}{\emph{ACM Computing Surveys (CSUR)}}
  \bibinfo{volume}{54}, \bibinfo{number}{6} (\bibinfo{year}{2021}),
  \bibinfo{pages}{1--25}.
\newblock


\bibitem[\protect\citeauthoryear{Parisi, Kemker, Part, Kanan, and
  Wermter}{Parisi et~al\mbox{.}}{2019}]%
        {parisi_continual_2019}
\bibfield{author}{\bibinfo{person}{German~I. Parisi}, \bibinfo{person}{Ronald
  Kemker}, \bibinfo{person}{Jose~L. Part}, \bibinfo{person}{Christopher Kanan},
  {and} \bibinfo{person}{Stefan Wermter}.} \bibinfo{year}{2019}\natexlab{}.
\newblock \showarticletitle{Continual {Lifelong} {Learning} with {Neural}
  {Networks}: {A} {Review}}.
\newblock \bibinfo{journal}{\emph{Neural Networks}}  \bibinfo{volume}{113}
  (\bibinfo{year}{2019}).
\newblock
\showISSN{08936080}
\urldef\tempurl%
\url{arxiv.org/abs/1802.07569}
\showURL{%
\tempurl}


\bibitem[\protect\citeauthoryear{Powers, Xing, Kolve, Mottaghi, and
  Gupta}{Powers et~al\mbox{.}}{2022}]%
        {powers2021cora}
\bibfield{author}{\bibinfo{person}{Sam Powers}, \bibinfo{person}{Eliot Xing},
  \bibinfo{person}{Eric Kolve}, \bibinfo{person}{Roozbeh Mottaghi}, {and}
  \bibinfo{person}{Abhinav Gupta}.} \bibinfo{year}{2022}\natexlab{}.
\newblock \showarticletitle{Cora: Benchmarks, baselines, and metrics as a
  platform for continual reinforcement learning agents}. In
  \bibinfo{booktitle}{\emph{Conference on Lifelong Learning Agents}}.
  Proceedings of Machine Learning Research.
\newblock


\bibitem[\protect\citeauthoryear{Raghavan, Hostetler, Sur, Rahman, and
  Divakaran}{Raghavan et~al\mbox{.}}{2020}]%
        {raghavan2020lifelong}
\bibfield{author}{\bibinfo{person}{Aswin Raghavan}, \bibinfo{person}{Jesse
  Hostetler}, \bibinfo{person}{Indranil Sur}, \bibinfo{person}{Abrar Rahman},
  {and} \bibinfo{person}{Ajay Divakaran}.} \bibinfo{year}{2020}\natexlab{}.
\newblock \showarticletitle{Lifelong learning using eigentasks: Task
  separation, skill acquisition, and selective transfer}.
\newblock \bibinfo{journal}{\emph{arXiv preprint arXiv:2007.06918}}
  (\bibinfo{year}{2020}).
\newblock


\bibitem[\protect\citeauthoryear{Rebuffi, Kolesnikov, Sperl, and
  Lampert}{Rebuffi et~al\mbox{.}}{2017}]%
        {rebuffi2017icarl}
\bibfield{author}{\bibinfo{person}{Sylvestre-Alvise Rebuffi},
  \bibinfo{person}{Alexander Kolesnikov}, \bibinfo{person}{Georg Sperl}, {and}
  \bibinfo{person}{Christoph~H Lampert}.} \bibinfo{year}{2017}\natexlab{}.
\newblock \showarticletitle{icarl: Incremental classifier and representation
  learning}. In \bibinfo{booktitle}{\emph{Proceedings of the IEEE conference on
  Computer Vision and Pattern Recognition}}. \bibinfo{pages}{2001--2010}.
\newblock


\bibitem[\protect\citeauthoryear{Riemer, Klinger, Bouneffouf, and
  Franceschini}{Riemer et~al\mbox{.}}{2019}]%
        {riemer2019scalable}
\bibfield{author}{\bibinfo{person}{Matthew Riemer}, \bibinfo{person}{Tim
  Klinger}, \bibinfo{person}{Djallel Bouneffouf}, {and}
  \bibinfo{person}{Michele Franceschini}.} \bibinfo{year}{2019}\natexlab{}.
\newblock \showarticletitle{Scalable recollections for continual lifelong
  learning}. In \bibinfo{booktitle}{\emph{Proceedings of the AAAI Conference on
  Artificial Intelligence}}, Vol.~\bibinfo{volume}{33}.
  \bibinfo{pages}{1352--1359}.
\newblock


\bibitem[\protect\citeauthoryear{Schaul, Quan, Antonoglou, and Silver}{Schaul
  et~al\mbox{.}}{2015}]%
        {schaul2015prioritized}
\bibfield{author}{\bibinfo{person}{Tom Schaul}, \bibinfo{person}{John Quan},
  \bibinfo{person}{Ioannis Antonoglou}, {and} \bibinfo{person}{David Silver}.}
  \bibinfo{year}{2015}\natexlab{}.
\newblock \showarticletitle{Prioritized experience replay}.
\newblock \bibinfo{journal}{\emph{arXiv preprint arXiv:1511.05952}}
  (\bibinfo{year}{2015}).
\newblock


\bibitem[\protect\citeauthoryear{Sculley, Holt, Golovin, Davydov, Phillips,
  Ebner, Chaudhary, Young, Crespo, and Dennison}{Sculley et~al\mbox{.}}{2015}]%
        {sculley2015hidden}
\bibfield{author}{\bibinfo{person}{David Sculley}, \bibinfo{person}{Gary Holt},
  \bibinfo{person}{Daniel Golovin}, \bibinfo{person}{Eugene Davydov},
  \bibinfo{person}{Todd Phillips}, \bibinfo{person}{Dietmar Ebner},
  \bibinfo{person}{Vinay Chaudhary}, \bibinfo{person}{Michael Young},
  \bibinfo{person}{Jean-Francois Crespo}, {and} \bibinfo{person}{Dan
  Dennison}.} \bibinfo{year}{2015}\natexlab{}.
\newblock \showarticletitle{Hidden technical debt in machine learning systems}.
\newblock \bibinfo{journal}{\emph{Advances in neural information processing
  systems}}  \bibinfo{volume}{28} (\bibinfo{year}{2015}).
\newblock


\bibitem[\protect\citeauthoryear{Simonyan and Zisserman}{Simonyan and
  Zisserman}{2014}]%
        {simonyan2014very}
\bibfield{author}{\bibinfo{person}{Karen Simonyan} {and}
  \bibinfo{person}{Andrew Zisserman}.} \bibinfo{year}{2014}\natexlab{}.
\newblock \showarticletitle{Very deep convolutional networks for large-scale
  image recognition}.
\newblock \bibinfo{journal}{\emph{arXiv preprint arXiv:1409.1556}}
  (\bibinfo{year}{2014}).
\newblock


\bibitem[\protect\citeauthoryear{Thrun and Mitchell}{Thrun and
  Mitchell}{1995}]%
        {thrun1995lifelong}
\bibfield{author}{\bibinfo{person}{Sebastian Thrun} {and}
  \bibinfo{person}{Tom~M Mitchell}.} \bibinfo{year}{1995}\natexlab{}.
\newblock \showarticletitle{Lifelong robot learning}.
\newblock \bibinfo{journal}{\emph{Robotics and autonomous systems}}
  \bibinfo{volume}{15}, \bibinfo{number}{1-2} (\bibinfo{year}{1995}),
  \bibinfo{pages}{25--46}.
\newblock


\bibitem[\protect\citeauthoryear{Vinyals, Ewalds, Bartunov, Georgiev,
  Vezhnevets, Yeo, Makhzani, K{\"u}ttler, Agapiou, Schrittwieser,
  et~al\mbox{.}}{Vinyals et~al\mbox{.}}{2017}]%
        {vinyals2017starcraft}
\bibfield{author}{\bibinfo{person}{Oriol Vinyals}, \bibinfo{person}{Timo
  Ewalds}, \bibinfo{person}{Sergey Bartunov}, \bibinfo{person}{Petko Georgiev},
  \bibinfo{person}{Alexander~Sasha Vezhnevets}, \bibinfo{person}{Michelle Yeo},
  \bibinfo{person}{Alireza Makhzani}, \bibinfo{person}{Heinrich K{\"u}ttler},
  \bibinfo{person}{John Agapiou}, \bibinfo{person}{Julian Schrittwieser},
  {et~al\mbox{.}}} \bibinfo{year}{2017}\natexlab{}.
\newblock \showarticletitle{Starcraft ii: A new challenge for reinforcement
  learning}.
\newblock \bibinfo{journal}{\emph{arXiv preprint arXiv:1708.04782}}
  (\bibinfo{year}{2017}).
\newblock


\bibitem[\protect\citeauthoryear{Wang, van~de Weijer, and Herranz}{Wang
  et~al\mbox{.}}{2021}]%
        {wang2021acae}
\bibfield{author}{\bibinfo{person}{Kai Wang}, \bibinfo{person}{Joost van~de
  Weijer}, {and} \bibinfo{person}{Luis Herranz}.}
  \bibinfo{year}{2021}\natexlab{}.
\newblock \showarticletitle{ACAE-REMIND for online continual learning with
  compressed feature replay}.
\newblock \bibinfo{journal}{\emph{Pattern Recognition Letters}}
  \bibinfo{volume}{150} (\bibinfo{year}{2021}), \bibinfo{pages}{122--129}.
\newblock


\bibitem[\protect\citeauthoryear{Yadan}{Yadan}{2019}]%
        {Yadan2019Hydra}
\bibfield{author}{\bibinfo{person}{Omry Yadan}.}
  \bibinfo{year}{2019}\natexlab{}.
\newblock \bibinfo{title}{Hydra - A framework for elegantly configuring complex
  applications}.
\newblock \bibinfo{howpublished}{Github}.
\newblock
\urldef\tempurl%
\url{https://github.com/facebookresearch/hydra}
\showURL{%
\tempurl}


\bibitem[\protect\citeauthoryear{Zbontar, Jing, Misra, LeCun, and Deny}{Zbontar
  et~al\mbox{.}}{2021}]%
        {zbontar2021barlow}
\bibfield{author}{\bibinfo{person}{Jure Zbontar}, \bibinfo{person}{Li Jing},
  \bibinfo{person}{Ishan Misra}, \bibinfo{person}{Yann LeCun}, {and}
  \bibinfo{person}{St{\'e}phane Deny}.} \bibinfo{year}{2021}\natexlab{}.
\newblock \showarticletitle{Barlow twins: Self-supervised learning via
  redundancy reduction}. In \bibinfo{booktitle}{\emph{International Conference
  on Machine Learning}}. PMLR, \bibinfo{pages}{12310--12320}.
\newblock


\bibitem[\protect\citeauthoryear{Zhang and Sutton}{Zhang and Sutton}{2017}]%
        {zhang2017deeper}
\bibfield{author}{\bibinfo{person}{Shangtong Zhang} {and}
  \bibinfo{person}{Richard~S Sutton}.} \bibinfo{year}{2017}\natexlab{}.
\newblock \showarticletitle{A deeper look at experience replay}.
\newblock \bibinfo{journal}{\emph{arXiv preprint arXiv:1712.01275}}
  (\bibinfo{year}{2017}).
\newblock


\end{thebibliography}


\end{document}